\documentclass[sigconf]{acmart}

\usepackage{pgfplots}
\usepackage{multirow}
\usepackage{graphicx}
\usepackage{soul}
\usepackage{url}
\usepackage[utf8]{inputenc}
\usepackage{graphicx}
\usepackage{amsmath}

\usepackage{bbding}
\usepackage{booktabs}
\usepackage{algorithm}
\usepackage{algorithmic}
\usepackage{bbm}

\usepackage{tikz}
\usetikzlibrary{automata}

\usepackage{xspace}
\usepackage{xcolor}
\usepackage{booktabs}
\usepackage{colortbl}
\usepackage{tabularx}
\usepackage{listings}
\usepackage{caption}
\usepackage{subcaption}

\usepackage{amsmath}
\usepackage{amsfonts}
\AtBeginDocument{%
  }

\newcommand{\toolname}{\textsc{MultiGain} 2.0}

\acmBadgeR[https://www.acm.org/publications/policies/artifact-review-and-badging-current]{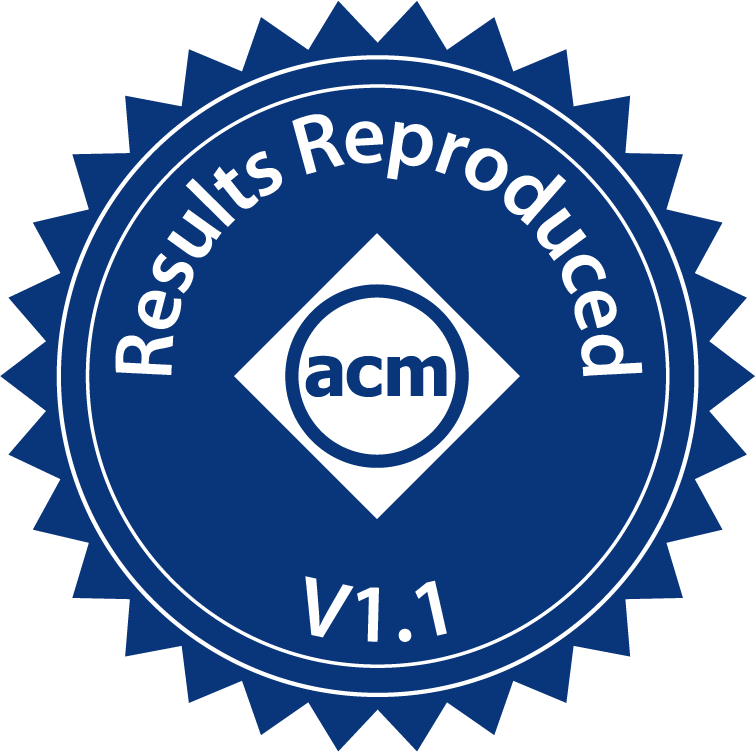}

\newcommand{\X}{{\ensuremath{\mathbf{X}}}}
\newcommand{\F}{{\ensuremath{\mathbf{F}}}}
\newcommand{\G}{{\ensuremath{\mathbf{G}}}}
\newcommand{\U}{{\ensuremath{\mathbf{U}}}}

\newcommand{\qee}{\hfill$\triangle$}

\newcommand{\para}[1]{\smallskip\noindent{\bf#1}}

\newcommand{\reward}{{r}}

\newcommand{\fcirc}[1]{\tikz\draw[#1,fill=#1] (0,0) circle (.7ex);}

\definecolor{bblue}{rgb}{0,0,1}

\definecolor{ggreen}{rgb}{0,0.5,0.1}

\definecolor{gb}{rgb}{0,0.8,1}

  \author{Severin Bals}
  \email{severin.bals@tum.de}
  \affiliation{%
  Technical University of Munich \country{Germany}
  }
  
  \author{Alexandros Evangelidis{ \Envelope}}
  \email{alexandros.evangelidis@tum.de}
  \affiliation{%
  Technical University of Munich \country{Germany}
  }
  \author{Jan K\v ret\'insk\'y}
  \email{jan.kretinsky@tum.de}
  \affiliation{%
  Technical University of Munich \country{Germany}
  }
  \affiliation{%
  Masaryk University Brno \country{Czech Republic}}

  \author{Jakob Waibel}
  \email{jakob.waibel@tum.de}
  \affiliation{%
  Technical University of Munich \country{Germany}
  }

\acmPrice{}
\copyrightyear{2024}
\acmYear{2024}
\setcopyright{rightsretained}
\acmConference[HSCC '24]{27th ACM International Conference on Hybrid Systems: Computation and Control}{May 14--16, 2024}{Hong Kong, Hong Kong}
\acmBooktitle{27th ACM International Conference on Hybrid Systems: Computation and Control (HSCC '24), May 14--16, 2024, Hong Kong, Hong Kong}
\acmDOI{10.1145/3641513.3650135}
\acmISBN{979-8-4007-0522-9/24/05}


\begin{document}

\title{MULTIGAIN 2.0: MDP controller synthesis for multiple mean-payoff, LTL and steady-state constraints}
\titlenote{This research was supported by the German Research Foundation (DFG) project 427755713 GOPro and the MUNI Award in Science and Humanities (MUNI/I/1757/2021) of the Grant Agency of Masaryk University.}



\begin{abstract}
We present \toolname, a major extension to the controller synthesis tool \textsc{MultiGain}, built on top of the probabilistic model checker PRISM. This new version extends \textsc{MultiGain}'s multi-objective capabilities, by allowing for the formal verification and synthesis of controllers for probabilistic systems with multi-dimensional long-run average reward structures, steady-state constraints, and linear temporal logic properties. Additionally, \toolname~ can modify the underlying linear program to prevent unbounded-memory and other unintuitive solutions and visualizes Pareto curves, in the two- and three-dimensional cases, to facilitate trade-off analysis in multi-objective scenarios.
\end{abstract}



\maketitle

\section{Introduction}
\label{sec:intro}

\para{Markov decision processes} (MDP), e.g.,~\cite{Puterman}, are the basic model for decision making in uncertain environments.
The policy synthesis problem is the problem of resolving the choices so that a given specification is satisfied.
In verification, there are many types of properties considered; in this work, we focus on \emph{infinite-horizon properties}.
Firstly, \emph{Linear Temporal Logic} (LTL) \cite{pnueli77} is mainstream in verification \cite{BK08}.
It can express complex temporal relationships, abstracting from the concrete quantitative timing, e.g., \emph{after every request, there is a grant} (not saying when exactly).
Secondly, \emph{Steady-State Policy Synthesis} (SS) \cite{akshay13} constrains the frequency with which states are visited, providing a more quantitative perspective.
Recently, it has started receiving more attention also in AI planning \cite{ijcai19,ijcai20,ltl-ss-jan}.
Thirdly, rewards provide a classic framework for quantitative properties.
In the setting of infinite horizon, a key role is played by the \emph{long-run average reward} (LRA, a.k.a.~mean payoff), e.g.,~\cite{Puterman}, which constrains the reward gained on average per step.
\begin{figure}[!t]
	\centering
			\begin{tikzpicture}[x=1.8cm,y=2cm,outer sep=1pt,thick, scale=0.7]
				\node[state,initial,initial text=] (s) at (0,0) {};
				\node[state] (t1) at (-1,-1) {};
				\node[state] (t2) at (-1,-2) {};
				{\node[state,fill=ggreen] (t21) at (-1,-2) {};}
				\node[state] (u1) at (0,-1) {};
				{\node[state,fill=ggreen] (u11) at (0,-1) {};}
				\node[state] (u2) at (0,-2) {};
				{\node[state,fill=red] (u21) at (0,-2) {};}
				\node[state] (v1) at (1,-1) {};
				\node[state] (v2) at (1,-2) {};
				{\node[state,fill=ggreen] (v21) at (1,-2) {};}
				\coordinate (x) at (1.75,-2);
				\path[->] 
				(s) edge node[left]{{0}} (t1)
				(t1) edge[bend right] node[left]{{\phantom{3}}{3}} (t2)
				(t2) edge[bend right] node[right]{{1}} (t1)
				(s) edge node[left]{{0}} (u1)
				(u1) edge[bend right] node[left]{{1}} (u2)
				(u2) edge[bend right] node[right]{{1}} (u1)
				(u2) edge[loop below] node[below]{{\phantom{3}}{3}} ()
				(s) edge node[left]{{0}} (v1)
				(v1) edge[] node[left]{{3}} (v2)
				(v2) edge[-] node[below]{{1}} (x)
				(x) edge[bend right=90pt,looseness=1] node[right]{{\small 0.5}} (v1)
				(x) edge[bend left=90pt,looseness=2] node[right]{{\small 0.5}} (v2)
				;
			\end{tikzpicture}
		\begin{itemize}

			\item linear temporal logic (LTL)
			$$\G(\fcirc{red}\implies\X(\neg \fcirc{red}\U\fcirc{ggreen}))$$
			
			\item steady-state constraints (SS)
			$$\fcirc{ggreen}\geq 0.6$$
			
			\item long-run average reward (LRA)
			$$\lim_{n\to\infty}(\inf) \frac 1n\sum_{i=1}^n \reward(A_i)$$
			
		\end{itemize}	
			
	\caption{An MDP and its heterogeneous specification}
	\label{fig:ltl}
		\end{figure}
\begin{example}
An example of an MDP with these specifications is shown in Fig.~\ref{fig:ltl}.
There is a non-trivial choice at the beginning, deciding, intuitively, in which set of states we shall be circulating forever. 
Such a set is called a maximal end component (MEC). 
Further, there is another choice in the middle MEC and a probabilistic transition in the right one.
The LTL formula in the example specifies that whenever a red state occurs, it is followed by non-red ones until a green one occurs.
This can be satisfied in all the MECs of this example; hence they are called \emph{accepting MECs}.
The steady-state constraint determines that we stay in green states at least 60\% of the time. 
Finally, the rewards are decorating the edges and then the average reward will be maximized on the ``red'' self-loop in the middle MEC. 
If all specifications are considered together, the reward is maximized in the right MEC only, because of the SS constraint. 
\qee

\noindent

\end{example}

We consider MDP with the LTL+SS+LRA specifications \emph{combining all these three types}, as introduced and theoretically solved in \cite{ltl-ss-jan}.
We build upon \textsc{MultiGain} \cite{multigain}, a tool extending PRISM \cite{prism} with multi-dimensional long-run average reward.
Our tool synthesizes a policy maximizing the LRA reward among all policies, ensuring the LTL specification (with the given probability) and adhering to the steady-state constraints. 

Our contribution can be summarized as follows:
\begin{itemize}
	\item 
	We extend \textsc{MultiGain} to analyze an MDP with a heterogeneous LTL+SS+LRA specification for maximizing the long-run average reward under the LTL and steady-state constraints, as described in \cite{ltl-ss-jan}.
	Additionally, we extend the specification and the algorithm to cater for further constraints. For example, satisfaction by policies that are deterministic (as in \cite{velasquez-2022}), unichain policies remaining in a single MEC, or policies with a bound on the size of their memory.
	\item
	We extend the syntax of the PRISM language slightly to accommodate the richer queries. 
	Further, we produce Pareto frontiers and display the two- and three-dimensional ones, to visualize the trade-offs.
	\item 
	We conduct a series of experiments to demonstrate the scalability of the tool.
\end{itemize}

\para{Related tools.} To the best of our knowledge, there are no tools that can simultaneously handle multi-dimensional LRA reward computation, LTL, and steady-state specifications. 
There are, however, two tools that handle multi-dimensional LRA objectives: (i) the previous version of \textsc{MultiGain} implements this functionality through linear programming which is also compatible with the solution offered in \cite{ltl-ss-jan} and implemented here; and (ii) STORM \cite{DBLP:conf/tacas/QuatmannK21}, which implements the same functionality more efficiently through value iteration. 
Additionally, the Partial Exploration Tool (PET) \cite{pet-tool} includes an implementation for LRA reward analysis, by focusing on partial exploration of the state space. 
However, it does not account for additional objectives such as LTL or steady-state specifications.
Furthermore, the work of \cite{velasquez-2022} presents a solution concept for finding deterministic \emph{unichain} policies under LTL and steady-state constraints, however, it does not include any reward structures.

\section{Functionality}
\label{sect:funct}
The main functionality of our tool is to answer multi-objective LRA queries constrained by LTL and steady-state specifications for MDPs and to synthesize a policy, if possible.
We begin with an overview of the tool's functionality, followed by a description of the various types of queries that are currently supported, including their syntax and semantics. Finally, we discuss additional functionalities that can be accessed via the command-line interface, and highlight key attributes of our tool.
The tool and supporting files for the results in the next section are available from \cite{files}.

\begin{center}
	\begin{figure*}[!t]
		\includegraphics[scale=0.6]{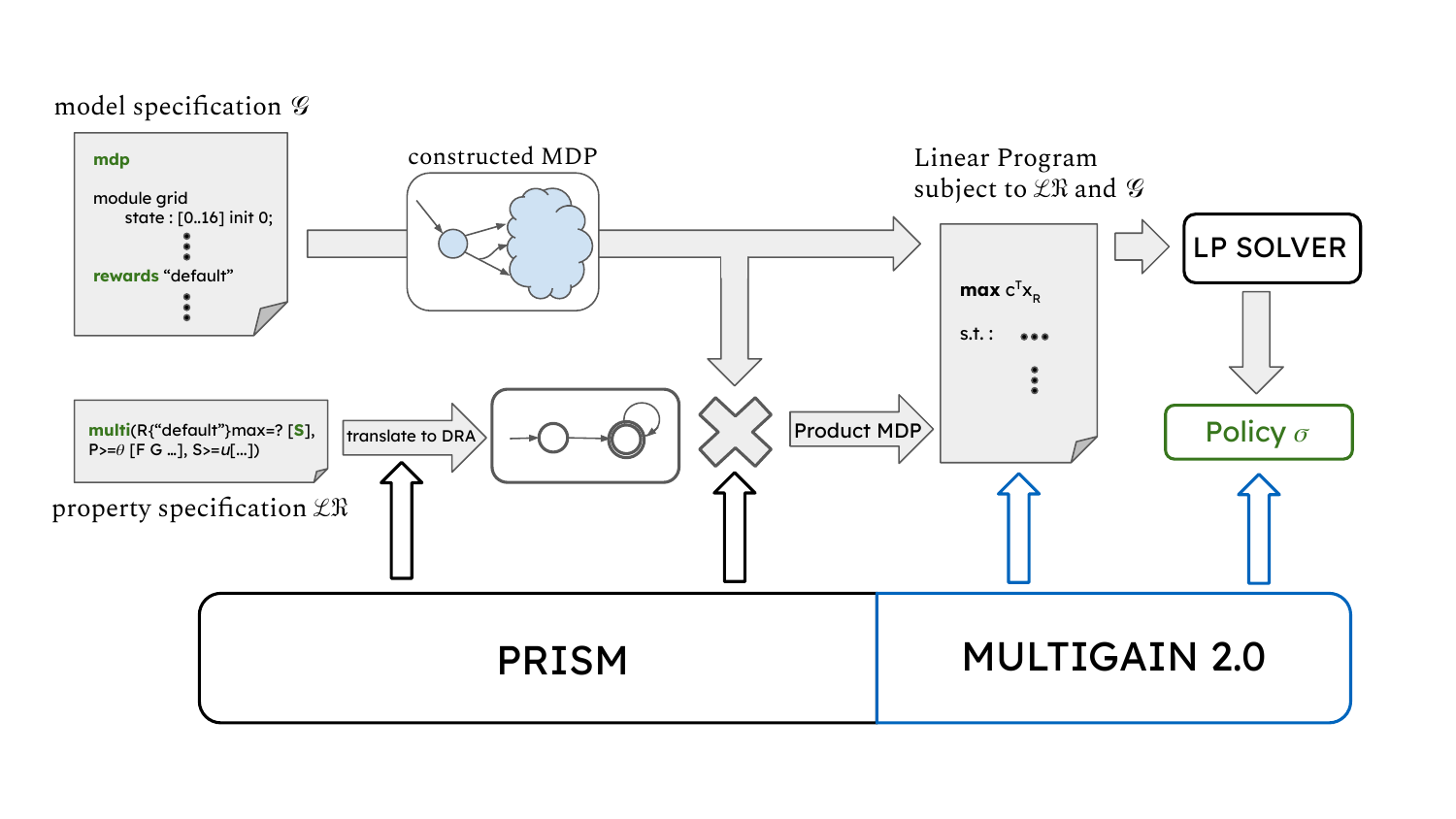}\vspace*{-3em}
		\caption{The workflow of \toolname~ from input specifications to output policy.}
		\label{workflow}
		
	\end{figure*}
\end{center}

\subsection{Workflow}
\label{sec:workflow}
Our tool functions according to the workflow depicted in Fig.~\ref{workflow}. 
The input consists of an MDP defined in the standard PRISM language\footnote{\href{https://www.prismmodelchecker.org/manual/ThePRISMLanguage/Introduction}{https://www.prismmodelchecker.org/manual/ThePRISMLanguage/Introduction}}, and an infinite-horizon property. 
This property is specified using an extension of PRISM's property specification
language\footnote{\href{https://www.prismmodelchecker.org/manual/PropertySpecification/Introduction}{https://www.prismmodelchecker.org/manual/PropertySpecification/Introduction}} that we developed.
PRISM starts by constructing the MDP from the input file and translates the specified LTL property into a Deterministic Rabin Automaton (DRA).
Then, it forms the product between the MDP and the DRA, also known as the \emph{product MDP}, which is then passed as an input to the novel component of our tool. 
Here, an LP is constructed, following the methodology described in \cite{ltl-ss-jan}, and fed to an LP solver.
Finally, after the LP is solved, \toolname~ extracts the solution from the solver and, if required, synthesizes a policy.

\def\code#1{\texttt{#1}}
\lstset{ %
	language=bash,                
	basicstyle=\footnotesize,       
	numbers=none,                   
	numberstyle=\footnotesize,      
	stepnumber=1,                   
	numbersep=5pt,                  
	backgroundcolor=\color{white},  
	showspaces=false,               
	showstringspaces=false,         
	showtabs=false,                 
	frame=single,           
	tabsize=2,          
	captionpos=b,           
	breaklines=true,        
	breakatwhitespace=true,    
	escapeinside={\%*}{*)}          
}
\lstset{postbreak=\raisebox{0ex}[0ex][0ex]
	{\ensuremath{\hookrightarrow\space}}}

\subsection{Example}
\label{sect:application}
\begin{figure}[]
	\centering
	\begin{subfigure}[b]{0.35\textwidth}
		\centering
		\includegraphics[scale=0.35]{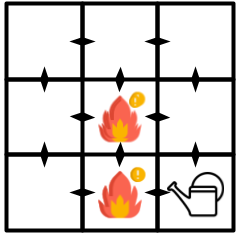}
		\subcaption{An example instance of the adjusted grid world model.}
		\label{fig:example_grid}
		\begin{minipage}{.9cm}
        \end{minipage}
	\end{subfigure}

	\begin{subfigure}[b]{\columnwidth}
		\centering
		\includegraphics[scale=0.38]{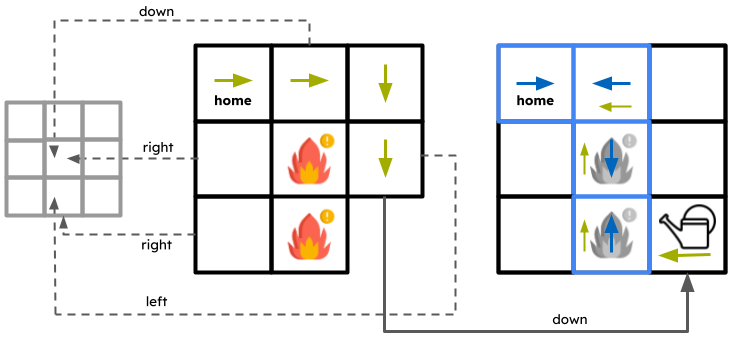}
		\subcaption{The computed product model and policy. The transient part of the policy is described with green arrows, the recurrent part with blue arrows. The blue framed state have positive switch probability from transient to recurrent behavior.}
		\label{fig:example_grid_policy}
	\end{subfigure}
	\caption{Example application of ~\toolname \ to an adjusted grid world model (a) and the corresponding solution (b).}
	\label{fig:examplepolicy}
\end{figure} 

Grid world models have been used extensively for the performance evaluation of various MDP algorithms and tools in fields such as reinforcement learning~\cite{rl-grid}, motion planning~\cite{plan-grid} and formal verification~\cite{velasquez-2022}. 
Here, we use two-dimensional grids of size $N \times N$, where an agent can traverse between the cells (or states) using one of the four actions, {\texttt{left}, \texttt{down}, \texttt{up}, \texttt{right}}, available in all states. 
Note that there are no actions to stay in a state i.e., no self loop actions. 
We show the grid world for $N=3$ in Fig.~\ref{fig:example_grid}. 
Additionally, two cells are labeled \texttt{danger} and one cell is labeled \texttt{water\_can} to indicate that they are on fire and the presence of a watering can, respectively.
The initial state, labeled \texttt{home} is the cell at the top left corner. Also, we define a reward structure, denoted as "extinguish", to assign a reward of 1 to the two fire states, intuitively encouraging the agent to repeatedly extinguish resurging flames.

An example query for our $3\times 3$ grid world with all three types of properties is shown below.
\begin{lstlisting}
	multi(R{"extinguish"}max=? [S], P>=1 [(! "danger") U "water_can"], S>=0.25 ["home"])
\end{lstlisting} 
Intuitively, it asks: \emph{``What is the maximum expected long-run average value of reward structure  "extinguish" (LRA), such that: (i) the agent does not visit any fire states before collecting the watering can first (LTL) and (ii) at least 25\% of the time in the long run it stays "home" (SS)? ''}. 

To compute a result for the query, as discussed in Section~\ref{sec:workflow} and shown in Fig.~\ref{fig:example_grid_policy}, the product MDP model is constructed, and the maximized LRA reward of $0.5$ is returned.
Intuitively, the product model consists of three copies of the grid, with a non-accepting MEC (gray grid on the left), which is reached when traversing to a fire cell before visiting the watering can, and the unique accepting MEC (right), reached when visiting the watering can cell first. 
The transient, recurrent and switching behavior of the policy is indicated by arrows in the product model in Fig. \ref{fig:example_grid_policy}. After the transient part directly guides around the fires to the watering can, it continues to traverse to any state with positive frequency in the long-run. At each such state the policy has a probability to switch to recurrent behavior, which suggests to loop both on the fire and the home cells. It may be noticed that while the four blue states have all positive occupation measure and are all reached by the policy, the recurrent behavior consists of two disconnected cycles here. 
Such unintuitive results can be avoided using further functionality of the tool.

\subsection{Infinite-horizon properties}
\label{sect:func-queries}

As described before, a multi-objective query for \toolname~ consists of the following specifications:
\begin{enumerate}
	\item \textbf{Long-run average}:
	Two types of LRA properties can be specified: (i) a \emph{numerical} property, which seeks to determine the maximum LRA achievable, or
	(ii) a \emph{Boolean} property that determines whether the LRA surpasses a certain threshold or not.
	In PRISM's syntax, a \emph{numerical} or a \emph{Boolean} LRA property could be represented as \code{R\{"rewardStruct"\}max=?[S]} or \code{R\{"rewardStruct"\} >=0.5[S]} respectively.
	
	\item \textbf{LTL}:
	The tool only supports a single Boolean query for LTL specifications, since multiple LTL formulae can be conjoined to form one formula.
	An example could look like \code{P>=0.75 [G F "stateLabel"]}, which expresses that with probability $\geq0.75$, states with the label \code{stateLabel} are reached infinitely often.
	
	\item \textbf{Steady-state}:
	An SS property of type \code{S<=0.1["stateLabel"]} requires that the steady-state probability distribution of the states with label \code{stateLabel} is bounded from above by $0.1$.
\end{enumerate}

\subsection{Syntax and semantics}
\label{sec:synt-sem}
As described in \cite{ltl-ss-jan}, there are several types of queries one can formulate by combining the properties discussed above.
Moreover, as previously discussed, we extended PRISM's syntax
to allow for new types of queries using the following notation:
\begin{equation*}
	{\rm keyword}\,\,\, ( [N,] prop_1, prop_2, \cdots, prop_k\code{)}
\end{equation*}
where the \code{keyword} can be one among \code{multi}, \code{mlessmulti}, \code{detmulti}, or \code{unichain} and each $prop_i$ is an LRA, LTL, or SS property. 
Note that there can be only one LTL property in the syntax of a query.
We now explain the semantics of the four different keywords.


\subsubsection{multi}
The semantics of the \code{multi} keyword is similar to its meaning in PRISM and \textsc{MultiGain}, i.e., computing a policy that satisfies the conjunction of all the individual properties.
Here, the type of result obtained depends on the number of numerical LRA properties.
If there are no numerical properties, the query only consists of Boolean LRA, LTL, and SS properties and the result is either \emph{true} or \emph{false}, depending on whether all of them can be satisfied or not.
 In the case of a single numerical LRA property, the tool returns the maximum (or minimum) achievable value for that LRA reward while satisfying all of the other properties. 
When more than one numerical property is given, the tool approximates the corresponding Pareto curve while satisfying all of the other properties.
An example of a \code{multi} query is shown below:
\begin{lstlisting}
	multi(R{"reward1"}max=? [S], R{"reward2"}max=? [S], R{"reward3"}>=0.5 [S], P>=0.75[G F "stateLabel1"], S>=0.5 ["stateLabel2"], S<=0.5["stateLabel2"])
\end{lstlisting}


\subsubsection{mlessmulti}
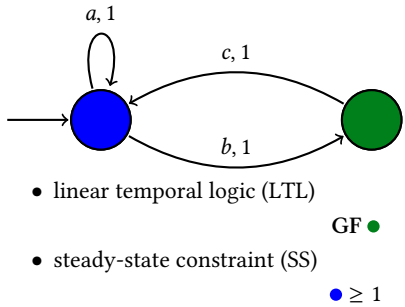
\begin{figure}[t]

\begin{minipage}{0.5\textwidth}
		\begin{tikzpicture}[x=1.8cm,y=2cm,outer sep=1pt,thick, scale=2]
			\node[state,initial,initial text=] (s) at (0,0) {};
			{\node[state,fill=bblue] (s1) at (0,0) {};}
			\node[state] (t) at (1,0) {};
			{\node[state,fill=ggreen] (t1) at (1,0) {};}
			\path[->] 
			(s) edge[loop above] node[above]{{$a,1$}} ()
			(s) edge[bend right] node[above]{{$b,1$}} (t)
			(t) edge[bend right] node[above]{{$c,1$}} (s)
			;
		\end{tikzpicture}
\end{minipage}
\begin{minipage}{0.5\textwidth}
	\begin{itemize}
		
		\item linear temporal logic (LTL)
		\begin{center}
		$\G \F$ \tikz\fill[ggreen] (0,0) circle (.6ex);
		\end{center}
		
		\item steady-state constraint (SS)
		\begin{center}
		\tikz\fill[bblue] (0,0) circle (.6ex); $\geq 1$
		\end{center}
		
	\end{itemize}	
		
\end{minipage}
\caption{An example MDP with LTL and steady-state constraints. Only policies visiting the accepting state less and less frequently satisfy both specifications, requiring unbounded memory for the information on a current run's history.}
\label{fig:unbound_memory}
\end{figure}

\begin{figure*}[!ht] 
    \centering
    \begin{subfigure}[b]{0.49\linewidth} 
        \centering
        \includegraphics[scale=0.23]{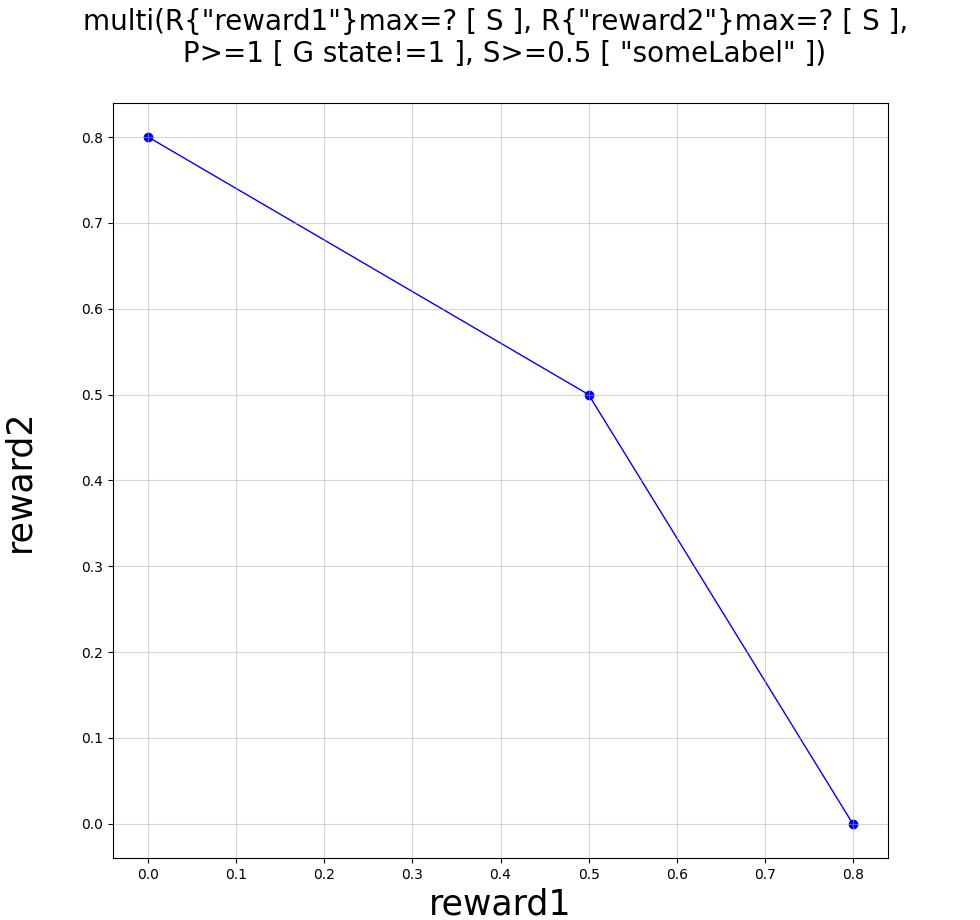}
        \caption{Approximated Pareto curve with 2 rewards.}
        \label{fig:2Dpareto}
    \end{subfigure}
    \hfill 
    \begin{subfigure}[b]{0.49\linewidth} 
        \centering
        \includegraphics[scale=0.17]{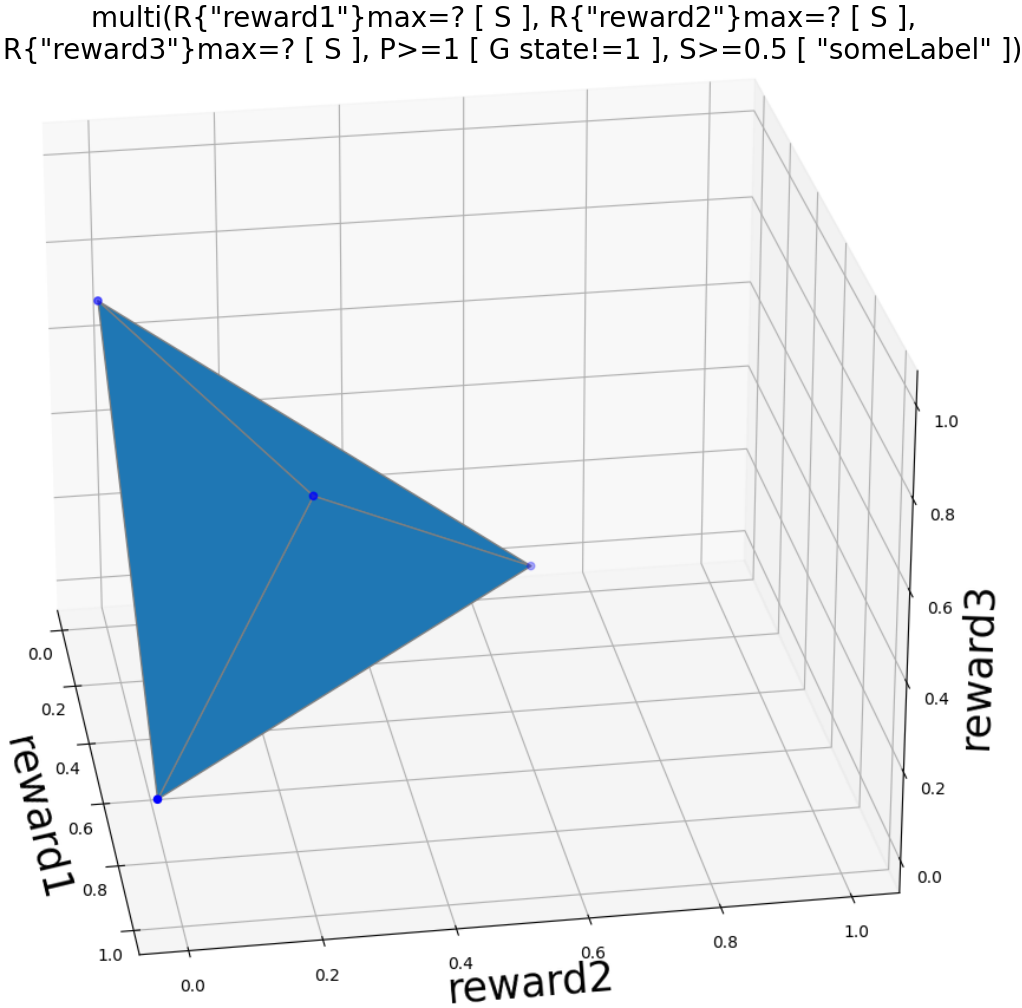}
        \caption{Approximated Pareto curve with 3 rewards.}
        \label{fig:3Dpareto}
    \end{subfigure}
    \caption{Example plots of approximated Pareto curves with 2 and 3 dimensions.}
    \label{fig:pareto}
\end{figure*}

The \code{mlessmulti} keyword, used in the previous version of our tool, solves a different problem in its current implementation.
In general, a policy computed by the LP (i.e., a \code{multi} query) might visit the accepting states less and less often to satisfy the
remaining constraints, thus requiring unbounded memory as seen in Fig. \ref{fig:unbound_memory}.
Relaxing the LRA and SS specifications by an arbitrary factor $\delta > 0$ lifts this restriction \cite{ltl-ss-jan},
such that a finite-memory policy exists for the model.

The implementation of a \code{mlessmulti} query addresses this problem from a different angle.
It allows the user to specify an additional integer $N$, signifying the maximum number of steps
on (the long-run) average before an accepting state is revisited.
The tool subsequently computes and outputs the
resulting minimal factor $\delta$ to uniformly relax all steady-state specifications and long-run average rewards,
with regards to this fixed accepting frequency. Hence, exporting the strategy yields a finite-memory policy,
more specifically a 2-memory policy \cite{ltl-ss-jan} consisting of a memoryless transient policy, which switches to a memoryless recurrent policy. Note that due to the modified objective function
it is not possible to define numerical LRA properties in a \code{mlessmulti} query.

An example of a \code{mlessmulti} query is shown below:
\begin{lstlisting}
	mlessmulti(1000, R>=0.5 ["sLabel"], P>=1[G F "tLabel"], S>=1 ["sLabel"])
\end{lstlisting} 

\subsubsection{detmulti}
Depending on the underlying model and property specification, the policy computed by the \code{multi} keyword may exhibit two
significant characteristics. Firstly, it is typically randomizing, and secondly, the policy may require an infinite amount of
memory to remember the current history. To address both of these issues, the \code{detmulti} keyword implements the approach by
\cite{velasquez-2022}, which is based on a mixed-integer linear program. The resulting policy, which is defined over the original
MDP rather than the product, is both deterministic and finite-memory.
The \code{detmulti} queries may contain a single LTL property and arbitrarily many steady-state specifications.
The result, other than an exportable policy, is either the optimal LRA reward or a boolean value indicating whether a solution was found or not.
An example of a \code{detmulti} query is shown below:
\begin{lstlisting}
	detmulti(P>=0.75 [(! "stateLabel1") U "stateLabel2"], S>=0.75 ["stateLabel3"])
\end{lstlisting} 

\subsubsection{unichain}
We introduce the keyword \code{unichain}, which computes a \emph{unichain} solution for the \code{multi} query, i.e.,
the recurrent behavior of the policy resides only in a single MEC and thus can be turned into a ``single'' behavior happening with probability 1.
Formally, a policy is called \emph{unichain} if the induced Markov chain has only one recurrent class and all the other states are transient. This is computed by exploring each MEC (or accepting MEC if an LTL specification is present) individually. The implementation concept follows the idea presented in \cite[Section 6]{ltl-ss-jan}.
If no numerical LRA properties are specified, the tool explores the MECs until a \emph{unichain} solution is found and outputs the corresponding boolean value. 
For a single numerical LRA property, our tool searches for the \emph{unichain} solution maximizing (or minimizing) the reward structure and outputs the corresponding reward.
Multiple numerical rewards are not allowed for this keyword, as this would result in comparing multiple Pareto curves. 
An example of a \code{unichain} query is shown below:

\begin{lstlisting}
	unichain((R{"reward1"}max=? [S], R{"reward2"}>=0.5 [S], P>=0.75[G F "stateLabel1"], S>=0.5 ["stateLabel2"], S<=0.5["stateLabel2"])
\end{lstlisting}


\subsection{Interface}
The tool is used via a command line interface, which requires the user to specify two files as input arguments,
containing the model and the queries.
The approximated Pareto curve can be exported to a file by using the flag -{}-\code{exportpareto}. 
Furthermore, the tool includes a Python script that enables the visualization of Pareto frontiers with two or three dimensions.
In Fig.~\ref{fig:pareto}, we show example plots of two- and three-dimensional Pareto frontiers produced by the tool. 
Moreover, for all queries except Pareto approximation, we provide the option to export the computed policy, which may have an unbounded memory, to a file in various formats.
%

%

\subsection{Implementation characteristics}
\label{sec:impl}
In this section, we report on the quality of \toolname~ by highlighting
some of its key characteristics.

\para{Extensibility.} The underlying LP solver implements a general interface and can thus be easily switched for every run.
This implementation allows the simple extension and addition of further LP solvers. 
	Currently, the tool supports the use of $\mathtt{lp\_solve}$ \cite{lp_solve} and $\mathtt{Gurobi}$ \cite{gurobi}.
After solving the LP, the tool extracts the solution from the solver and, if required, synthesizes a policy.

For approximating Pareto curves a new generic class has been implemented which takes as input a weight function, mapping weights to reward structures. 
This class lifts the Pareto curve approximation from \toolname~ so that other PRISM-based tools could utilize it. 
Furthermore, Pareto curves of any dimension can be approximated, contrary to the two-dimensional limit of the previous version of the tool.
Since the tool is implemented in the unifying approach of the PRISM pipeline, it can be extended at a variety of entry points, as seen in Fig.~\ref{workflow}. 
For example, new deterministic automata could be implemented alongside the translation of the LTL and building the product model, without changing the tool's core functionality. 

\section{Experimental Evaluation}\label{experimental_eval}
In this section, we assess the performance of our tool in terms of its ability to solve the types of queries described in Section~\ref{sec:synt-sem}.
We conducted multiple experiments to evaluate the performance of our tool.
We first discuss the experimental setup, followed by the technical details regarding our experiments, and then we give a detailed overview of our experimental results in Section~\ref{sec:exp:overview}.

\para{Experimental setup.} Our evaluation consists of three parts: (i) the evaluation of the full property suite (LRA, SS, and LTL properties) using a grid world model;
(ii) a scalability analysis of the tool regarding its performance with an increasing number of steady-state constraints; and (iii) an evaluation of how different LP solvers
impact the tool's efficiency, including both runtime performance and memory usage, in the context of handling queries. 
%
\\
\para{Technical details.} All experiments were performed on a desktop computer with 16 GB of RAM and
an Intel i7-8550U CPU @ 1.80GHz,
running Ubuntu 22.04.3 LTS. 

For the grid world model, the average running time over 20 runs was recorded, as a countermeasure to the high variance of individual running times.
All results are rounded to three decimal places.
\begin{table}[!t]
    \caption{Average running time (in seconds) over 20 randomly grid world labeled instances.}
    \label{gridtable}
    \setlength{\tabcolsep}{2pt}
    \centering
\small
        \begin{tabular}{l c | c c c c c}
            & LRA & \multicolumn{5}{c}{Average running time for each grid} \\ 
            LTL  & $\mathtt{R}^{\mathit{rew\_c}}_{\mathtt{max}=?}[\,\mathtt{S}\,]$  & $4\times4$ & $16\times16$ & $32\times32$ & $64\times64$ & $128\times128$ \\[0.5ex] 
            \hline
            \hline
            $\G (\neg b) \land (\G \F a)$ & $\times$ & $0.121$ & $0.231$ & $0.466$ & $1.296$ & $26.104$ \\ 
            \hline
            $(\G \F a) \lor (\F \G b)$ & $\times$ & $0.029$ & $0.091$ & $0.210$ & $0.581$ & $2.498$\\
            \hline
            $(\F a) $\U $ b$ & $ \times$ & $0.020$ & $0.074$ & $0.238$ & $0.852$ & $4.042$ \\
            \hline
            $(\F a) \land (\F b) \land (\F c)$ & $ \times$ & $0.042$ & $0.147$ & $0.499$ & $2.396$ & $21.678$ \\ 
            \hline
            \hline
            $\G (\neg b) \land (\G \F a)$ & \checkmark & $0.025$ & $0.092$ & $0.623$ & $10.12$ & $128.197$ \\
            \hline
            $(\G \F a) \lor (\F \G b)$ & \checkmark & $0.021$ & $0.081$ & $0.555$ & $5.245$ & $106.772$\\
            \hline
            $(\F a) $\U $ b$ &  \checkmark & $0.013$ & $0.105$ & $0.668$ & $11.053$ & $158.448$ \\
            \hline
            $(\F a) \land (\F b) \land (\F c)$ & \checkmark  & $0.017$ & $0.298$ & $3.232$ & $81.389$ & $883.446$\\ [0.5ex]
        \end{tabular}

\end{table}

\subsection{Results}
\label{sec:exp:overview}
\para{LRA+LTL+SS queries.}
In Table~\ref{gridtable}, we present the results for various \code{multi} queries that involve the combination of all three types of properties. 
These queries are categorized into two groups: those containing an LRA property, denoted by a \checkmark symbol, and those that do not, represented by a $\times$ symbol. 
Following the experiments in~\cite{velasquez-2022}, the states were randomly divided into four equally-sized subsets, and each subset was labeled with an atomic proposition from the set $AP = {a, b, c, d}$ before a run of the tool. 
Additionally, we created a reward structure, $rew\_c$, that assigns a reward of 1 to each state labeled with $c$. 
In each run, we required the LTL formula to be fulfilled with a probability threshold of $\theta = 0.5$ and steady-state constraints of $S \geq 0.01 [$"$d$"$]$, $S \leq 0.5 [$"$d$"$]$.

The upper half of Table~\ref{gridtable} highlights the efficient performance of our tool when no LRA property is specified.
Even for the largest grid world MDP model, the longest running time is still only a few seconds. 
Also, when an LRA property is specified (lower half of Table~\ref{gridtable}), for the majority of the cases in the $64\times64$  grid, it remains quite efficient ($<11$ seconds);
however, scalability issues start emerging for the $128\times 128$ grid.
The LRA property used here maximizes the average reward of $rew\_c$, denoted as $\mathtt{R}^{\mathit{rew\_c}}_{\mathtt{max}=?}[\,\mathtt{S}\,]$.
For larger grid sizes, such as the $128\times 128$ case, the runtimes are reasonable except for the last LTL property $(\F a) \land (\F b) \land (\F c)$. 
This trend aligns with the findings in \cite{velasquez-2022} and can be attributed to the random generation procedure of the grid world instances, which may have a bias toward certain types of models.

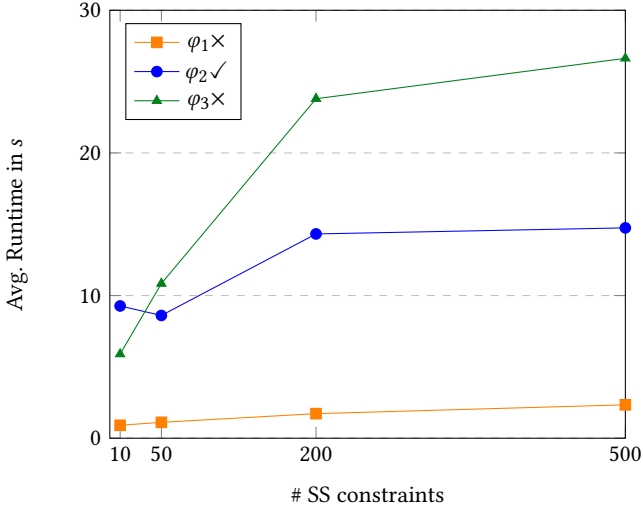
\begin{figure}[!t]
    \centering
    \begin{tikzpicture}
\begin{axis}[
    title={},
    xlabel={\# SS constraints},
    ylabel={Avg. Runtime in $s$},
    xmin=0, xmax=500,
    ymin=0, ymax=30,
    xtick={10,50,200,500},
    ytick={},
    legend pos=north west,
    ymajorgrids=true,
    grid style=dashed,
]
\addplot[color=orange,mark=square*]
    coordinates {
  (10,0.9063)(50,1.113)(200,1.721)(500,2.34)
  };
\addplot[color=blue,mark=*]
    coordinates {
   (10,9.27)(50,8.605)(200,14.320)(500,14.745)
   };
\addplot[color=ggreen,mark=triangle*]
    coordinates {
    (10,5.897)(50,10.84)(200,23.798)(500,26.63)
    };
\legend{$\varphi_1 \times$, $\varphi_2$\checkmark, $\varphi_3 \times$}
\end{axis}
\end{tikzpicture}
    \caption{LP solver running times of \code{multi} queries on $64\times64$ grid world models with the LTL formulae $\varphi_1, \varphi_2$ and $\varphi_3$ from above, based on the number of steady-state constraints specified. 
    }
    \label{fig:scaling_ss}
\end{figure}
\para{Scaling the number of steady-state constraints.}
In the next set of experiments, we systematically assess the computational overhead as a function of the number of
steady-state constraints introduced per specification. We consider the following three different LTL formulae: $\varphi_1 = (\G \F a) \vee (\F \G b)$, $\varphi_2 = (\F a) \U b$, and $\varphi_3 = \F a \wedge \F b \wedge \F c$, 
and we employ instances of a $64\times64$ grid world for our experimental setup. 
To instantiate non-trivial steady-state constraints, for each experimental iteration, we stochastically select a subset of states, 
denoted $\mathcal{S}$, that are labeled with $d$. Every state $s \in \mathcal{S}$ is then attributed a distinct label, denoted $l_s$.
Formally, for a given label $l_s$, its corresponding constraint is such that:
${U}(l_s) = 0.5$, and $\sum_{s \in \mathcal{S}} {L}(l_s)  \leq 0.25$, where $U$ and $L$ denote an upper and lower bound, respectively.
For each of these designated labels $l_s$, we append a steady-state constraint to the property specification,
ensuring that the cumulative lower bounds do not exceed a threshold, thereby reducing the likelihood of encountering infeasible scenarios.

Our experiments span configurations with 10, 50, 200, and 500 steady-state constraints, evaluated against the three distinct
LTL formulae.
To account for the variance in individual running times, as discussed in previous experiments, the recorded running times were averaged over 20 runs.
We note that for $\varphi_1$, the solver's average runtime increases as more steady-state constraints are appended. 
Specifically, starting from an average runtime of $\approx 1$ second with 10 constraints, it rose to 2.34 seconds with 500 constraints.
On the other hand, $\varphi_2$ presented an interesting pattern as the average runtime of $\approx 9$ seconds
for 10 and 50 constraints, were fairly similar. However, a significant increase was observed as we introduced 200 constraints,
reaching $\approx$ 14 seconds, and this growth seemed to stabilize by the time we integrated 500 constraints.
Finally, $\varphi_3$ seemed to be the most computationally demanding, starting at $\approx 6$ seconds with 10 constraints and
reaching $\approx 27$ seconds at 500 constraints.

This experiment demonstrates that adding steady-state constraints does not have a
significant impact on overall runtime, and is therefore not a restriction on the user.
\begin{table} [!t]
	\caption{Average LP solver runtimes (in seconds) over 20 runs of respective grid world instances, recorded using $\mathtt{Gurobi}$ and $\mathtt{lp\_solve}$. The faster runtime of each problem is marked in green.
}
	\label{table_grid_lpvsgurobi} 
	\centering
	\footnotesize 
	\begin{tabular}{|c|c|c|p{0.5cm}|p{0.7cm}|p{0.7cm}|p{0.78cm}|}
		
		\hline
		\multirow{2}{*}{$\mathtt{R}^{\mathit{rew\_c}}_{\mathtt{max}=?}[\,\mathtt{S}\,]$ (LRA)} & & & \multicolumn{4}{c|}{\textbf{Average running time per grid size}} \\
		& LTL & Solver & $4\times4$ & $16\times16$ & $32\times32$ & $64\times64$\\
		\hline
		\multirow{6}{*}{$\times$} & \multirow{2}{*}{$\varphi_1$} & $\mathtt{Gurobi}$ & $0.002$ & \textcolor{ggreen}{$0.029$} & \textcolor{ggreen}{$0.062$} &  \textcolor{ggreen}{$0.264$}\\
		
		& & $\mathtt{lp\_solve}$ & \textcolor{ggreen}{$0.001$} & $0.072$ & $0.33$ & $1.165$\\\cline{2-7}

		& \multirow{2}{*}{$\varphi_2$} & $\mathtt{Gurobi}$ & $0.007$ &\textcolor{ggreen} {$0.018$} & \textcolor{ggreen}{$0.09$} & \textcolor{ggreen}{$0.405$}\\

		& & $\mathtt{lp\_solve}$ &\textcolor{ggreen} {$0.001$} & $0.069$ & $1.502$ & $1.617$\\\cline{2-7}
		
		& \multirow{2}{*}{$\varphi_3$} & $\mathtt{Gurobi}$ & $0.002$ & \textcolor{ggreen}{$0.027$} & \textcolor{ggreen}{$0.233$} & \textcolor{ggreen}{$2.006$}\\

		& & $\mathtt{lp\_solve}$ & \textcolor{ggreen}{$0.001$} & $0.097$ & $1.369$ & $15.224$\\\cline{1-7}
		
		\multirow{6}{*}{\checkmark} & \multirow{2}{*}{$\varphi_1$} &$\mathtt{Gurobi}$ & $0.006$ & $\textcolor{ggreen}{0.038}$ & $0.342$ & $4.566$\\
		
		& & $\mathtt{lp\_solve}$ & \textcolor{ggreen}{$0.002$} & $0.059$ & \textcolor{ggreen}{$0.206$} & \textcolor{ggreen}{$1.856$}\\\cline{2-7}
		
		& \multirow{2}{*}{$\varphi_2$} & $\mathtt{Gurobi}$ & $0.005$ &  $0.097$ & \textcolor{ggreen}{$0.284$} & $10.022$\\

		& & $\mathtt{lp\_solve}$ & \textcolor{ggreen}{$0.002$} & \textcolor{ggreen}{$0.071$} & $0.412$ & \textcolor{ggreen}{$3.842$}\\\cline{2-7}

		& \multirow{2}{*}{$\varphi_3$} & $\mathtt{Gurobi}$ & $0.003$ & \textcolor{ggreen}{$0.014$} & \textcolor{ggreen}{$2.311$} & \textcolor{ggreen}{$79.566$}\\
		
		& & $\mathtt{lp\_solve}$ & \textcolor{ggreen}{$0.001$} & $0.595$ & $3.043$ & $160.526$ \\
		
		\hline
	\end{tabular}
	
\end{table}

\para{LP solver comparison.}
In this set of experiments, we evaluate how the performance of the queries is affected by the choice of the underlying LP solver.
We consider the three LTL formulae used in the previous experiments.
As stated in Section~\ref{sec:impl}, \toolname~ supports the publicly available solver $\mathtt{lp\_solve}$ as well as the well-known
commercial state-of-the-art solver $\mathtt{Gurobi}$.
In Table~\ref{table_grid_lpvsgurobi} we present the average runtime over 20 runs of $\mathtt{Gurobi}$ and $\mathtt{lp\_solve}$
on various grid world instances, with and without LRA maximization, marked by the $\checkmark$ and $\times$ symbols, respectively.
The fastest runtime for each problem is marked in green. 

Our results show a trend where the $\mathtt{Gurobi}$ significantly outperforms $\mathtt{lp\_solve}$ as the size of the grid increases,
especially on the $64 \times 64$ grid. However, even in this case, there are instances in which $\mathtt{lp\_solve}$ performs reasonably well.
For example, with the LTL formula $\varphi_1$, when LRA reward maximization is considered, $\mathtt{lp\_solve}$'s
 average runtime of $1.856$ seconds outperforms $\mathtt{Gurobi}$'s of $\approx 4.5$ seconds.
Moreover, for the $\varphi_3$ formula under the same grid configuration and with LRA maximization included, $\mathtt{Gurobi}$ is $\approx 50\%$ faster compared to
$\mathtt{lp\_solve}$, whereas in the case without LRA maximization, $\mathtt{Gurobi}$ demonstrates $\approx 87\%$ reduction in runtime over $\mathtt{lp\_solve}$.

On the other hand, in smaller grid sizes, $\mathtt{lp\_solve}$ becomes more competitive and in some cases it even outperforms $\mathtt{Gurobi}$.
For instance, when evaluating the $\varphi_1$ formula without LRA maximization on a $4 \times 4$ grid, $\mathtt{lp\_solve}$ achieves a runtime
of $0.001$ seconds compared to $\mathtt{Gurobi}$'s $0.002$ seconds, denoting a $50\%$
more efficiency in handling smaller grid sizes.

Apart from the runtime performance of both solvers, we also compared the quality of the solution on selected experiments.
Both solvers exhibited no noticeable issues in terms of finding a solution and solution quality.

\begin{figure}[]
		\includegraphics[scale=0.3]{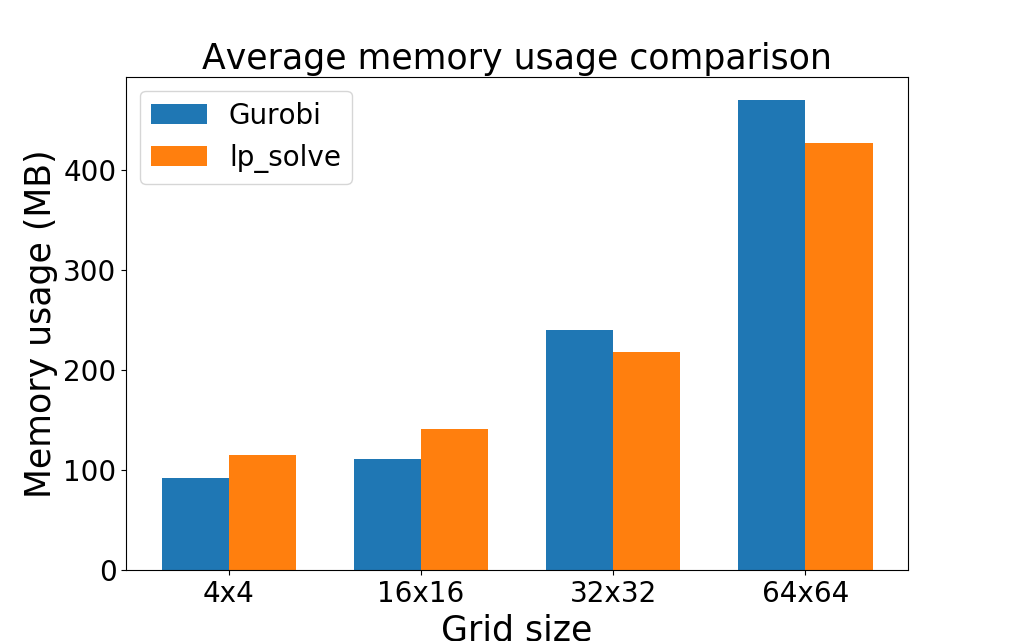}
		\caption{Average memory usage (in megabytes) over 20 runs of respective grid world instances, recorded using $\mathtt{Gurobi}$ and $\mathtt{lp\_solve}$.}
		\label{mem}
		
	\end{figure}

In Fig.~\ref{mem} we report on the average memory usage of our tool when using the $\mathtt{Gurobi}$ and $\mathtt{lp\_solve}$ solvers,
across different grid sizes.
As expected, there is an increasing trend in terms of the tool's memory usage as the grid size increases, indicating greater memory requirements
for solving larger problems.
We note, that for the majority of the cases, the tool consumes more memory when using $\mathtt{lp\_solve}$ compared to $\mathtt{Gurobi}$, 
something which can be observed in smaller grid sizes.
However, across every grid size the difference in memory usage between the two solvers is not significant, indicating that
both solvers are viable alternatives from a memory consumption perspective.

\section{Conclusion}
\label{sect:conclusion}
%

We presented \toolname, an MDP controller synthesis tool for multiple
long-run average reward structures subject to LTL and steady-state constraints.
Apart from the normal combination of these different objectives, it is also able to solve the $\delta$-satisfaction problem, which relaxes the objectives by a small factor $\delta$.
We also implemented a new method sketched in \cite{ltl-ss-jan} providing \emph{unichain} solutions, and a method described in \cite{velasquez-2022} for \emph{deterministic} solutions.
Our tool can export the Pareto curve and the policy, and it can also visualize two and three-dimensional Pareto curves.

This tool can be further extended to work for \emph{omega-regular} objectives.
Another useful direction for future work would be to explore combinations with other types of properties, such as non-linear,
\emph{finite-horizon} or \emph{discounted}, rewards. 

\begin{acks}
The authors would like to thank Ismail R. Alkhouri for the comprehensive guidance on their approach in \cite{velasquez-2022} and provision of example models, as well as Ayse Aybüke Ulusarslan for their initial help with the project.
Further, the authors would like to thank the original developers of \textsc{MultiGain} for allowing us to use the tool's name for our extension.

\end{acks}

\bibliographystyle{ACM-Reference-Format}
\bibliography{ref,ijcai21}





\end{document}